\documentclass[conference]{IEEEtran}
\IEEEoverridecommandlockouts
\usepackage{epsfig}
\usepackage{times}
\usepackage{amsmath}
\usepackage{amssymb}
\usepackage[numbers]{natbib}
\usepackage{multicol}
\usepackage[bookmarks=true]{hyperref}
\usepackage{multirow}
\usepackage[T1]{fontenc}  
\usepackage[utf8]{inputenc}
\usepackage[english]{babel}
\usepackage[export]{adjustbox}
\usepackage{graphicx}
\usepackage{blindtext}
\usepackage{siunitx}
\usepackage{booktabs, threeparttable}
\usepackage{subcaption}
\usepackage{tikz}
\usepackage{tikzscale}
\usepackage{pgfplots}
\usepackage{layouts}
\graphicspath{{graphics/}}
\usepackage[colorinlistoftodos, german]{todonotes}
\usepackage[T1]{fontenc}
\usepackage[utf8]{inputenc}

\newcommand{\set}[1]{\mathcal{#1}}

\newcommand{\lowerBound}[1]{{#1}^-}
\newcommand{\transpose}[1]{{#1}^{\intercal}}
\newcommand{\horizon}{{T}}
\newcommand{\samplingStepWidth}{T_{s}}
\newcommand{\weight}{{w}}
\newcommand{\positionCartesianX}{{x}}
\newcommand{\positionCartesianY}{{y}}
\newcommand{\positionCartesianXY}{\mathbf{x}}
\newcommand{\positionFrenetLon}{{s}}  %

\newcommand{\maneuver}{{\mathfrak{m}}}
\newcommand{\maneuverSet}{{\set{M}}} %
\definecolor{blue}{rgb}{0, 0.5, 0.8}
\definecolor{paleblue}{rgb}{0.5, 0.7, 1}
\definecolor{red}{rgb}{0.82, 0.1, 0.2}
\definecolor{green}{rgb}{0, 0.5, 0.0}
\definecolor{orange}{rgb}{1, 0.7, 0.0}

\usepackage{hyperref}
\hypersetup{
    colorlinks=true,
    linkcolor=black,
    citecolor=black,
    filecolor=black,
    urlcolor=black,
}
\title{
Tackling Existence Probabilities of Objects with \\ Motion Planning for Automated Urban Driving}
\author{\authorblockN{}
\authorblockN{Ömer Şahin Taş and Christoph Stiller}
\authorblockA{FZI Research Center for Information Technology \& 
Karlsruhe Institute of Technology\\
Karlsruhe, Germany\\
tas@fzi.de}}
\usepackage{textcomp}
\usepackage[absolute,showboxes]{textpos}
\newcommand*{\myfont}{\fontfamily{lmss}\selectfont} %
\textblockrulecolour{white}
\newcommand{\presentationinformation}{
\begin{textblock}{14.0}(0.45,0.25)    %
    \vspace{2mm}
    \noindent{\bfseries{\myfont{\footnotesize{Presented at the Interaction and Decision-Making in Autonomous-Driving workshop \\
    at the Robotics: Science and Systems (RSS) conference, July 13th 2020.}}}}
\end{textblock}
}
\begin{document}
\presentationinformation
\maketitle
\thispagestyle{empty}
\pagestyle{empty}
\begin{abstract} 

Motion planners take uncertain information about the environment as an input. 
The environment information is often quite noisy and
has a tendency to contain false positive object detection.
State-of-the-art motion planners consider all objects alike, thus producing overcautious behavior.
In this paper we present a planning approach that considers alternative maneuvers in a combined fashion and plans a motion that is formed by the probabilities of those alternatives.
The proposed planner can smoothly react to objects with low existence probability while remaining collision-free in case their existence substantiates.
In this way, it tolerates the faults arising from perception and prediction, thus reducing their impact on operational reliability.
\end{abstract}\IEEEpeerreviewmaketitle

\section{Introduction}
\label{sec:introduction}

Automated driving needs to employ various sensor modalities to meet the requirements for driving.
The redundant sensor information is processed in different modules and 
the divergence in the subsequent measurements is typically represented probabilistically.
The resulting data is then fused 
either by a temporal filtering intrinsically, e.g.\ \cite{Duraisamy2015},
or as a single shot with subsequent temporal filtering, e.g.\ in grid maps \cite{Richter2019}, \cite{Steyer2018}.
In any case, the fusion may generate false negatives,
and more frequently false positive objects in order to avoid any severe consequences. 
The output of the fusion is then processed in scene understanding module and 
the predicted motion of the objects is transmitted to the motion planner \cite{tas2017automated}.
A motion planner receives all the objects and considers the uncertainties associated with them
by treating these as hard or soft constraints.

\begin{figure}[t!]
\vspace{2mm}
\includegraphics[width=\columnwidth]{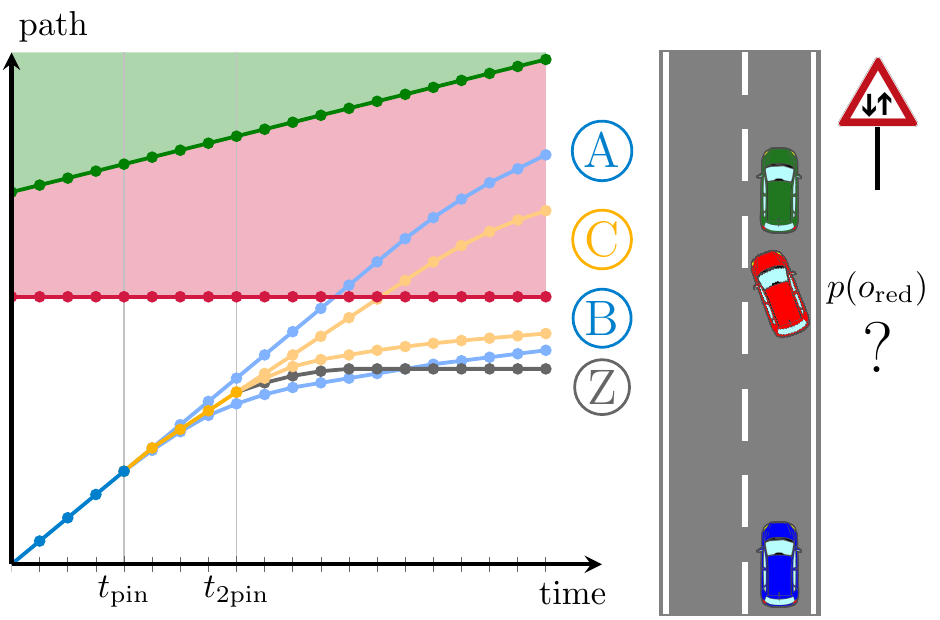}
\caption{A scenario in which a phantom object (the red car) with existence probability $p (o_\mathrm{red})$ suddenly appears. 
A conventional planner has two alternative plans to execute:
{Plan A}; which ignores the phantom object,
Plan B; which treats the phantom object as a real object.
Our planner considers the phantom object according to its existence probability resulting in Plan C together with Plan Z; which is its safe evasive maneuver.
Because of continuous replanning, only $t \in (t_\mathrm{pin}, t_{2\mathrm{pin}}]$ will be executed,
the rest will be replanned.}
\label{fig:concept}
\end{figure}

The object list transmitted to the planner may contain 
a false-positive object, i.e.\ a \textit{phantom} object, close to the ego vehicle (see Fig.~\ref{fig:concept}).
Independent of the existence probability of the phantom object,
this will trigger either braking or swerving on a multi-lane road.
On the other hand, 
the low existence probability of the phantom may diminish completely within the next series of perception updates.
This typically corresponds to durations smaller than half a second \cite{yin2008real}, \cite{kocamaz2019}.
In such, braking or swerving will be discarded resulting in unstable behavior.

Tackling phantom objects can be done in different layers of an automated vehicle. 
This can be done either by a module that is closely related to the perception or fusion, e.g.\ \cite{barnes2017find}
or by a centralized plausibility module, e.g.\ \cite{tas2017automated}.
Another approach is to consider the problem from the end-side: 
\textit{how urgent is it for the motion planner to react to the object}?
The answer to this question depends on both \textit{the existence probability of the object}, and
\textit{its pose and prediction uncertainty}.
Our motion planning approach fundamentally considers these two aspects by incorporating their probabilities in
planning and thus tolerates faulty detection until it receives more precise inputs over time.
It further ensures collision-free motion even if the situation with low probability becomes more probable.
In this sense, we tackle the existence probability of objects from the motion planning perspective for the first time.

The rest of the paper is structured as follows: 
we classify the uncertainty types in Section~\ref{sec:uncertainty},
since understanding the source and types of uncertainties that propagate in an automated vehicle is essential.
Then, in Section~\ref{sec:related_work},
we present the state-of-the-art solvers that use distinct modalities to deal with those uncertainties and 
describe their limits. 
Our motion planner has certain input requirements on the predicted object list.
In Section~\ref{sec:prediction}, we briefly describe these requirements, 
followed by Section~\ref{sec:planning},
where we elaborate on the problem formulation to tackle existence probabilities in planning.

Lastly, Section~\ref{sec:conclusion} concludes the paper and summarizes the key aspects of our research.

\section{Types of Uncertainties in Automated Driving}
\label{sec:uncertainty}

The uncertainties are either due to noisy measurements, or the obscurity of the future, or the occlusions.
In a robotic system, these uncertainties yield to three main classes of problems: 
\begin{enumerate}
\item Uncertainty in pose
\item Uncertainty in prediction
  \begin{enumerate}
  \item Route
  \item Maneuver intention
  \item Profile
  \end{enumerate}
\item Uncertainty in existence
  \begin{enumerate}
  \item Field-of-view \textit(fov)
  \item Phantom detections.
  \end{enumerate}
\end{enumerate}

The \textit{uncertainty in pose} covers the position and velocity uncertainty of the ego vehicle and other participants.
This type of uncertainty is typically represented with a probability distribution, e.g.\ normal distribution.

The second class, the \textit{uncertainty in prediction}, comprises the route alternatives, intentions creating homotopic maneuver classes, and their motion profiles.
Representing this type of uncertainty is quite difficult as it is highly related with types of objects, e.g.\ 
pedestrians, bicycles, motorbikes, trucks, unclassified etc. 
Each one of these types introduces different maneuver classes, depending on the current traffic scene and the interactions.
In unstructured environments or for some object types in structured environments, 
a semantic classification of maneuvers is not even possible.
It must be underlined that the uncertainty in pose is covered by this type of uncertainty while planning motion over a horizon.

The \textit{uncertainty in existence} is the type of uncertainty, 
which this paper focuses on.
It reflects the uncertainty of an object to exist, 
either at the outside borders of fov, or even inside the fov.
Following the notion presented in \cite{tas2018limited}, 
we refer to the objects at the border of fov as \textit{hypothetical} objects.
However, we refer to any object suspected to be a non-object as a
phantom\footnote{We prefer using ``phantom'' over ``ghost'' as this word etymologically originates from the ancient Greek word \textit{ph\'antasis} which is more concerned with appearance and perception.} object.
We set the existence probability of objects to be constant over the planning horizon.

\section{Existing Planners that Consider Uncertainty}
\label{sec:related_work}

Planning safe maneuvers while considering uncertainty is a profound topic of motion planning research for automated driving.

Motivated from robotics applications, 
early automated vehicle motion planning methods penalized the probability of being in the space of other objects while planning a path \cite{xu2014motion}.
Safe intersection crossing with probabilistic risk indicators is tackled in \cite{de2014collision}.
Recent works have focused on occlusions and considered the objects that could emerge behind the visible field.
Orzechowski et al.\ used reachable sets \cite{althoff2016set} for dealing with occlusions in a safe way \cite{orzechowski2018tackling}.
Other works utilized sampling based approaches to deal with the same problem:
samples can either be used for estimating the risk associated with occlusions and hence allowing to plan smooth motion \cite{yu2019occlusion},
or creating probabilistic motion plans \cite{naumann2019safe}.
Tas and Stiller calculated the reachability of an object whose state is modeled with a normal distribution.
They solve the resulting optimization problem while ensuring the point-of-no-return to satisfy safety margins in occlusions \cite{tas2018limited}.

Gritschneider et al.\ focused on interaction uncertainties in a Markov Decision Process (MDP) and
chose the high-level behavior actions that are sent to the motion planner \cite{gritschneder2016adaptive}.
Among distinct maneuver options, the planner considered undecided cases as well. 
But it failed to consider the uncertainties in state and maneuver intentions of other vehicles.
Zhan et al.\ used the planning approach developed by Ziegler et al.\ \cite{ziegler2014trajectory},
and proposed to plan maneuvers that are optimal for undecided cases \cite{zhan2016non}.
In their paper, they did not specify the optimization method 
or provide any detail on computation time.
Apart from this work, Tas et al.\ used the same approach to deal with the same problem \cite{tas2018decision}.
In their paper, they planned for all maneuver options including the undecided case.
Considering the entropy of predicted maneuver intention of the other vehicle, they
executed the best maneuver that satisfies a threshold on entropy value.

Partially Observable MDP (POMDP) methods consider the state as a probability distribution and perform predictions on the motion of other vehicles while planning over a horizon.
Although these are very well suited for motion planning problems of automated vehicles,
they are very complex to solve in real-time \cite{papadimitriouComplexityMarkovDecision1987}.
Even though early applications in automated driving scenarios were scene-specific \cite{brechtel2014probabilistic},
recent sampling-based solvers are able to tackle scenario-independent problems in real time without any training \cite{egorov2017pomdps}.
Hubmann et al.\ use POMDPs to deal with all types of uncertainties except the existence uncertainty of phantom objects \cite{hubmann2019pomdp}.
Based on our definitions, they use the term ``phantom object'' to tackle hypothetic objects that 
could be at the boundaries of the visible field.
Another recent work focuses on uncertainties resulting from occlusions, limited sensor range, probabilistic prediction and unknown intentions by using model predictive control \cite{sun2019behavior}.
They used inverse reinforcement learning to fit the cost function to observed maneuvers.

\section{Requirements on Predicted Outputs}
\label{sec:prediction}

The planner requires predicted motion of other participants over the planning horizon and 
the existence of these objects as an input.
As outlined in Section~\ref{sec:uncertainty}, 
an object has uncertainties regarding
type information, existence probability, maneuver classes, and the profile of these classes.

The profile is required to be represented as truncated normal distribution.
The reason for choosing normal distribution is its ease of computation,
eventually causing the problem to remain quadratic.
The mean value of the distribution can be obtained from an arbitrary prediction algorithm, 
either model based or learning based \cite{lefevre2014survey}.
The calculation of variance is not straightforward due to the constraints employed by a truncation.
An overview is provided in \cite{simon2010kalman}.
Together with type information, 
every maneuver class creates a modus in a multi-modal truncated normal distribution.

\section{Tackling Uncertainties for Safe Planning}
\label{sec:planning}

We solve the motion planning problem for a kinematic vehicle model by transforming it into an optimization problem \cite{ziegler2014trajectory}.
Our previous work has already developed the approach presented in \cite{ziegler2014trajectory} further
by integrating safe stops while considering perception uncertainties, limited visible field and uncompliant behaviours \cite{tas2018limited}.
Our follow-up work has resolved situations in which the intention of other vehicles are unclear
by analyzing \emph{undecided} cases about the maneuver intention of others 
and executing undecided motion plans \cite{tas2018decision}.
Here we build upon these by tackling phantom detections.
We briefly present the fundamentals from those papers and 
elaborate these for handling phantom objects.

\subsection{Baseline}
\label{sec:planning_baseline}

For a motion to be optimal, we seek smooth control inputs.
This can be achieved by penalizing jumps in acceleration values of a maneuver $\maneuver \in \maneuverSet$ over a planning horizon $\horizon$.
A discrete representation of such a functional with stepwidth $\samplingStepWidth$ by $N$ points
using forward differences minimizes the sum
\begin{equation} \label{eq:cost_function}
J^\mathrm{d} (\positionCartesianXY_0, \positionCartesianXY_1, \ldots, \positionCartesianXY_{N-1}) = \sum_{i=0}^{N-4} L (\positionCartesianXY_i, \positionCartesianXY_i^\mathrm{d}, \positionCartesianXY_i^\mathrm{dd}, \positionCartesianXY_i^\mathrm{ddd} ). 
\end{equation}
The variable ${\positionCartesianXY}_i = \transpose{[\positionCartesianX_i, \positionCartesianY_i]}$ represents $i$th motion support point 
corresponding to the position values in Cartesian coordinates,
$L$ is a function comprising cost terms minimizing \textit{value} or \textit{range} residuals 
(see \cite{tas2018decision} and \cite{tas2016making} for details),
the superscript `$\mathrm{d}$' indicates that the variable is a discretized approximation.

\begin{figure}[h!]
    \vspace{2mm}
    \centering
    \captionsetup[subfigure]{width=\textwidth}
    
    \vspace{1mm}
    \begin{subfigure}[t]{0.98\columnwidth}
        \includegraphics[width=\columnwidth]{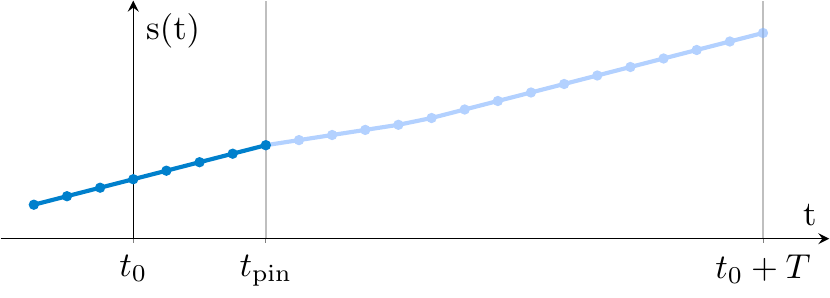} 
        \caption{The opaque part of the motion is taken from the previous solution and the transparent part is altered during planning.}
        \label{fig:planning_base} 
    \end{subfigure}
    
    \vspace{1mm}
    \begin{subfigure}[t]{0.98\columnwidth}
        \includegraphics[width=\columnwidth]{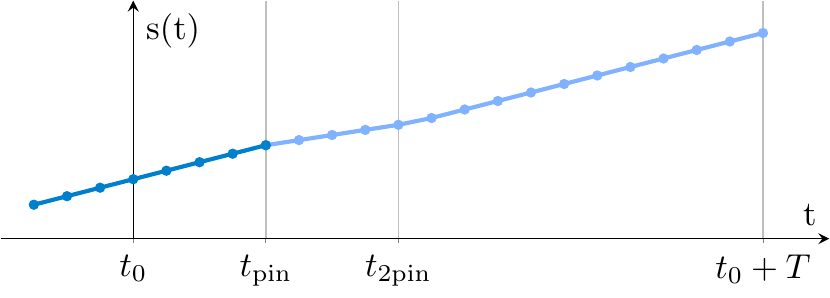} 
        \caption{The part of the motion that will be executed is in the interval $t \in (t_\mathrm{pin}, t_{2\mathrm{pin}}]$.}
        \label{fig:planning_exec} 
    \end{subfigure}

   \vspace{1mm}  
    \begin{subfigure}[t]{0.98\columnwidth}
        \includegraphics[width=\columnwidth]{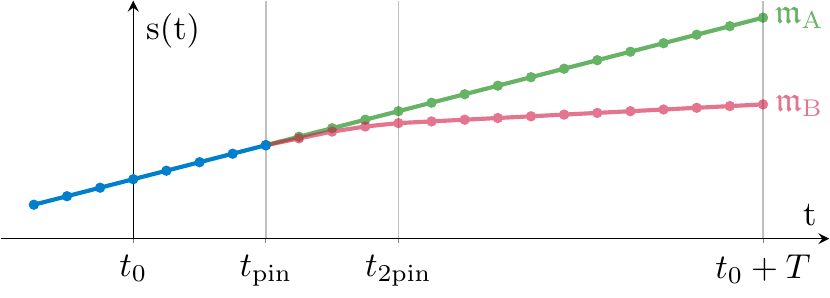} 
        \caption{Standard approaches calculate the homotopic maneuver alternatives $\maneuver_\mathrm{A}$ and $\maneuver_\mathrm{B}$ separately.}
        \label{fig:two_independent_variants}
    \end{subfigure}
    
    \vspace{1mm}
    \begin{subfigure}[t]{0.98\columnwidth}
        \includegraphics[width=\columnwidth]{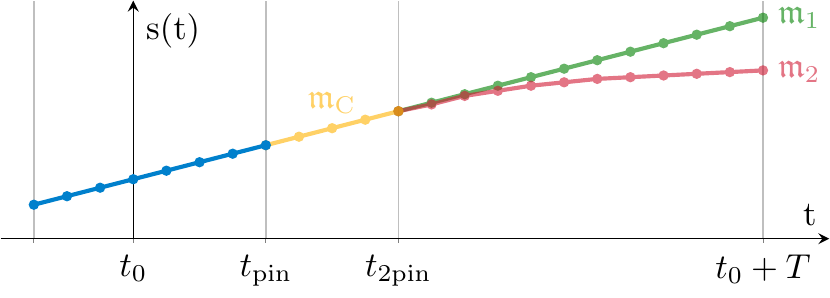} 
        \caption{Undecided motion of maneuver $\maneuver_\mathrm{C}$ obtained by calculating $\maneuver_\mathrm{A}$ and $\maneuver_\mathrm{B}$ with the same weight in the interval  $t \in (t_\mathrm{pin}, t_{2\mathrm{pin}}]$.}
        \label{fig:two_timeshift_variants}
    \end{subfigure}
    
    \caption{Path-time diagram of a planned trajectory at time $t_0$.}
    \label{fig:pairwise_subfig}
\end{figure}

The modules of an automated vehicle have delays and the motion is not planned instantaneously.
Therefore, to maintain temporal consistency during replanning, 
some motion support points are taken from the previously planned motion.
We denote the index until which the previous motion is kept fixed with the subscript `$\mathrm{pin}$'.
The visualization is given in Fig.~\ref{fig:planning_base},
after the time $t_\mathrm{pin}$, a motion is calculated.
If replanning is performed in constant time intervals,
only the part between the time interval $t \in (t_\mathrm{pin}, t_{2\mathrm{pin}}]$
of the motion planned at $t_0$ will be executed.
The support points between $t \in [t_{2\mathrm{pin}}, \horizon)$ is replanned during the next planning instance,
i.e.\ at $t_{2\mathrm{pin}}$ (see Fig.~\ref{fig:planning_exec}).

Until now, the motion for a single maneuver $\maneuver$ is considered.
An automated vehicle typically has to consider maneuvers of different homotopy classes and choose one of them
according to utility and safety reserves \cite{bender2015combinatorial}.
Even though a variety of maneuver alternatives are possible, the promising maneuvers in most cases reduce to two, 
i.e.\ \textit{pass} or \textit{yield}.
We denote two maneuvers of different homotopy classes with $\mathfrak{m}_\mathrm{A}$ and $\mathfrak{m}_\mathrm{B}$.
Almost every motion planning algorithm plans these alternatives separately (see Fig.~\ref{fig:two_independent_variants}). 

Another option is to plan these maneuvers in a combined fashion \cite{zhan2016non}, \cite{tas2018decision}.
In this way, the optimization parameter vector becomes
\begin{align}
\mathbf{X} &= 
\nonumber
(\underbrace{\positionCartesianXY_{0}, \ldots \positionCartesianXY_\mathrm{pin}}_{\text{pinned}}, \, \underbrace{\positionCartesianXY_{\mathrm{pin}+1}, \ldots \positionCartesianXY_{2\mathrm{pin}}}_{\text{shared} \, (\mathfrak{m}_\mathrm{C}) }, \, \\
& \underbrace{\positionCartesianXY_{2\mathrm{pin} + 1}, \ldots \positionCartesianXY_{N}}_{\mathfrak{m}_1}, \, \underbrace{\positionCartesianXY_{N+1}, \ldots \positionCartesianXY_{2N - n_\mathrm{pin}}}_{\mathfrak{m}_2}).
\end{align}
This allows for consideration of both homotopy classes simultaneously and results in a motion that is tailored for undecided cases:
the motion of maneuver $\mathfrak{m}_\mathrm{C}$ is equally optimal for $\mathfrak{m}_\mathrm{A}$ and $\mathfrak{m}_\mathrm{B}$ (see Fig.~\ref{fig:two_timeshift_variants}).
The implementation details for constructing constraints and calculating the cost are provided in \cite{tas2018decision}.
Although this approach is advantageous for considering uncertainties, 
such a computation has an increased complexity:
$2N - 2 n_\mathrm{pin}$ number of free variables are optimized instead of $N - n_\mathrm{pin}$,
and the matrices of the minimization problem are not diagonal anymore.

\subsection{Maintaining Safety}
\label{sec:planning_safety}

A set of rules for an automated vehicle to plan a safe motion is provided in Responsibility Sensitive Safety model (RSS) \cite{shalevshwartz2017formal}.
In continuous replanning, we consider a planned motion \textit{safe} if
the vehicle is able to reach a safe state after executing the motion that will be kept fixed in the next planning timestep.
In the notation presented in the previous subsection,
the motion in  $t \in (t_\mathrm{pin}, t_{2\mathrm{pin}}]$ must ensure the presence of safe maneuvers.

These safe maneuvers can either be full braking actions, or swerve maneuvers, or a combination of them.
For urban environments, we treat full braking actions $\mathfrak{m}_\mathrm{Z}$ as safe maneuvers and ensure the presence of them
by introducing an inequality constraint.
The full stop point $\positionFrenetLon_\mathrm{stop}$ along the path must remain lower than 
the lower bound of the truncated normal distribution representing the position of the closest object $o$ 
for timesteps
\begin{equation}
\lowerBound{o}_i - \positionFrenetLon_\mathrm{stop} \geq 0 \,\,, \forall i \in \{ n_\mathrm{pin}, \ldots , 2 n_\mathrm{pin} -1 \}.
\end{equation}

In case of limited visibility, we consider hypothetic objects at the end of the visible field as described in \cite{tas2018limited}.
Unless a vehicle is percepted, we treat the hypothetic vehicle to be uncompliant to the traffic rules, 
and approach the intersection while reserving a full braking maneuver that comes to a full stop before this zone.
Once a vehicle is percepted, if its yield intention becomes clear by reducing its speed so that it can come to full stop before colliding,
we deactivate the constraint applied on the planner.

\subsection{Maintaining Comfort}
\label{sec:planning_chance}

All of the motion support points have an influence on comfort.
The change in lateral and longitudinal acceleration is considered as a cost summand to yield a comfortable ride \cite{ziegler2014trajectory}.
Besides this, we soft constrain the collision uncertainty with the objects by calculating cumulative distribution function (cdf).
We take truncated normal distributed position predictions (see \cite{tas2018decision}).
The error function (erf) is used in the calculation of cdf.
We use an efficient numerical approximation of the erf that is presented in \cite[p.\ 214]{press1992numerical}.

\subsection{Interacting with Unknown Maneuver Intentions}
\label{sec:safe_interact}

Our planner considers two maneuver alternatives while interacting with other traffic participants.
This number can be increased at the cost of solving problems in a parallel way.
In some cases, the maneuver intention of others is unclear and it is beneficial to perform a neutral
maneuver, as described in Section~\ref{sec:planning_baseline}.
Our approach presented in \cite{tas2018decision} selected the maneuver yielding the lowest cost
among the alternatives that have an entropy lower than a certain threshold.

Note that such interactions can also be considered as a further comfort term,
because they estimate the further evolution of the current situation and 
subsequently serve for adapting the speed.

\subsection{Tackling Existence Probabilities}
\label{sec:planning_existence}

Objects inside the object list sent to the planner must be processed carefully.
Independent of the value of existence probability, 
the safety conditions provided in the Section~\ref{sec:planning_safety} must be met.
If no immediate evasive action is required, we follow a strategy derived from the undecided case.

The situation depicted in Fig.~\ref{fig:concept}
resembles a situation in which the existence of a detected vehicle is unclear.
The planner has again two maneuver alternatives:
$\maneuver_\mathrm{A}$ and $\maneuver_\mathrm{B}$.
The red vehicle which is a potential phantom is away, i.e.\
$\positionFrenetLon_\mathrm{B} (t_\mathrm{pin}) > \positionFrenetLon_\mathrm{stop} (t_\mathrm{pin})$,
such that no immediate reaction $\maneuver_\mathrm{Z}$ is required.
The vehicle can execute an undecided maneuver following the approach presented in Section~\ref{sec:planning_baseline}.

An undecided motion can be planned by considering detection probabilities.
$p (o_\mathrm{red})$ denotes the existence probability of the red vehicle,
whereas $p^{\mathrm{TP}}$, $p^{\mathrm{FP}}$, $p^{\mathrm{TN}}$, $p^{\mathrm{FN}}$
denote true positive, false positive, true negative, and false negative detection probabilities of the detector, respectively.
The weights of the two maneuver alternatives $\maneuver_\mathrm{A}$ and $\maneuver_\mathrm{B}$ can be obtained by
\begin{subequations}
\begin{align}
\weight_\mathrm{A} &= (1 - p (o_\mathrm{red})) \, p^{\mathrm{TN}} +   p (o_\mathrm{red}) \, p^{\mathrm{FP}} \\
\weight_\mathrm{B} &= p (o_\mathrm{red}) \, p^{\mathrm{TP}} + (1 - p (o_\mathrm{red})) \, p^{\mathrm{FN}} .
\end{align}
\end{subequations}
The resulting motion is obtained by weighting cost terms $J_\mathrm{A}$ and $J_\mathrm{B}$ with $\weight_\mathrm{A}$ and $\weight_\mathrm{B}$
\begin{align}
\nonumber
J^\mathrm{d} & (\positionCartesianXY_0, \positionCartesianXY_1, \ldots, \positionCartesianXY_{2N-n_\mathrm{pin}}) = \weight_\mathrm{A} \, J^\mathrm{d}_\mathrm{A} \, (\positionCartesianXY_{0}, \ldots \, , \positionCartesianXY_{N}) \, + \\
& \weight_\mathrm{B} \, J^\mathrm{d}_\mathrm{B} \, (\positionCartesianXY_{0}, \, \ldots \, ,  \positionCartesianXY_{2\mathrm{pin}}, \, \positionCartesianXY_{N+1}, \, \ldots \, , \positionCartesianXY_{2N - n_\mathrm{pin}} ). 
\end{align}
In this way, the planner reacts to the object smoothly, without initiating harsh braking maneuvers.

\section{Conclusions and Future Work}
\label{sec:conclusion}

The existing literature focuses on developing new motion planning algorithms with different modalities.
However, the biggest challenge here is to tolerate the faults in perception.
While the literature has overseen this need,
our research focused on developing a motion planner to compensate the faults in perception.

In this paper, we categorized types of uncertainties influencing the motion of an automated vehicle.
We clarified the reasons of these uncertainties and defined the underlying causes.
Based on this, we considered existence probabilities of objects in planning
by inspecting the motion reserve to activate evasive actions.
This allowed us to tolerate faulty detections of phantom objects.
With our previous work, we were able to present a motion planner 
that could deal with all types of uncertainties while ensuring safety. 

\bibliographystyle{plainnat}
\fontsize{9.5pt}{11pt} \selectfont
\bibliography{paper}

\begin{thebibliography}{29}
\providecommand{\natexlab}[1]{#1}
\providecommand{\url}[1]{\texttt{#1}}
\expandafter\ifx\csname urlstyle\endcsname\relax
  \providecommand{\doi}[1]{doi: #1}\else
  \providecommand{\doi}{doi: \begingroup \urlstyle{rm}\Url}\fi

\bibitem[Althoff and Magdici(2016)]{althoff2016set}
Matthias Althoff and Silvia Magdici.
\newblock Set-based prediction of traffic participants on arbitrary road
  networks.
\newblock \emph{IEEE Trans.\ Intell.\ Veh.}, 1\penalty0 (2):\penalty0 187--202,
  2016.

\bibitem[Barnes et~al.(2017)Barnes, Maddern, and Posner]{barnes2017find}
Dan Barnes, Will Maddern, and Ingmar Posner.
\newblock {Find Your Own Way: Weakly-Supervised Segmentation of Path Proposals
  for Urban Autonomy}.
\newblock In \emph{Proc.\ IEEE Int.\ Conf.\ Robot.\ and Autom.}, pages
  203--210, 2017.

\bibitem[Bender et~al.(2015)Bender, Ta{\c{s}}, Ziegler, and
  Stiller]{bender2015combinatorial}
Philipp Bender, {\"O}mer~{\c{S}}ahin Ta{\c{s}}, Julius Ziegler, and Christoph
  Stiller.
\newblock The combinatorial aspect of motion planning: Maneuver variants in
  structured environments.
\newblock In \emph{Proc.\ IEEE Intell.\ Veh.\ Symp.}, pages 1386--1392, 2015.

\bibitem[Brechtel et~al.(2014)Brechtel, Gindele, and
  Dillmann]{brechtel2014probabilistic}
Sebastian Brechtel, Tobias Gindele, and R{\"u}diger Dillmann.
\newblock Probabilistic decision-making under uncertainty for autonomous
  driving using continuous pomdps.
\newblock In \emph{Proc.\ IEEE Intell.\ Trans.\ Syst.\ Conf.}, pages 392--399,
  2014.

\bibitem[de~Campos et~al.(2014)de~Campos, Runarsson, Granum, Falcone, and
  Alenljung]{de2014collision}
Gabriel~R de~Campos, Adam~H Runarsson, Fredrik Granum, Paolo Falcone, and Klas
  Alenljung.
\newblock Collision avoidance at intersections: A probabilistic
  threat-assessment and decision-making system for safety interventions.
\newblock In \emph{Proc.\ IEEE Intell.\ Trans.\ Syst.\ Conf.}, pages 649--654,
  2014.

\bibitem[Duraisamy et~al.(2015)Duraisamy, Schwarz, and Wohler]{Duraisamy2015}
Bharanidhar Duraisamy, Tilo Schwarz, and Christian Wohler.
\newblock {On track-to-track data association for automotive sensor fusion}.
\newblock \emph{Int.\ Conf.\ Inform.\ Fusion}, 2015.

\bibitem[Egorov et~al.(2017)Egorov, Sunberg, Balaban, Wheeler, Gupta, and
  Kochenderfer]{egorov2017pomdps}
Maxim Egorov, Zachary~N Sunberg, Edward Balaban, Tim~A Wheeler, Jayesh~K Gupta,
  and Mykel~J Kochenderfer.
\newblock {POMDPs. jl: A framework for sequential decision making under
  uncertainty}.
\newblock \emph{The J.\ of Mach.\ Learn.\ Res.}, 18\penalty0 (1):\penalty0
  831--835, 2017.

\bibitem[Gritschneder et~al.(2016)Gritschneder, Hatzelmann, Thom, Kunz, and
  Dietmayer]{gritschneder2016adaptive}
Franz Gritschneder, Patrick Hatzelmann, Markus Thom, Felix Kunz, and Klaus
  Dietmayer.
\newblock Adaptive learning based on guided exploration for decision making at
  roundabouts.
\newblock In \emph{Proc.\ IEEE Intell.\ Veh.\ Symp.}, pages 433--440, 2016.

\bibitem[Hubmann et~al.(2019)Hubmann, Quetschlich, Schulz, Bernhard, Althoff,
  and Stiller]{hubmann2019pomdp}
Constantin Hubmann, Nils Quetschlich, Jens Schulz, Julian Bernhard, Daniel
  Althoff, and Christoph Stiller.
\newblock A pomdp maneuver planner for occlusions in urban scenarios.
\newblock In \emph{Proc.\ IEEE Intell.\ Veh.\ Symp.}, pages 2172--2179, 2019.

\bibitem[Kocamaz(2019)]{kocamaz2019}
Mehmet Kocamaz.
\newblock Drive labs: Tracking objects with surround camera vision, Jun 2019.
\newblock URL
  \url{https://news.developer.nvidia.com/drive-labs-tracking-objects-with-surround-camera-vision/}.
\newblock {Date retrieved: January 10, 2020}.

\bibitem[Lef{\`e}vre et~al.(2014)Lef{\`e}vre, Vasquez, and
  Laugier]{lefevre2014survey}
St{\'e}phanie Lef{\`e}vre, Dizan Vasquez, and Christian Laugier.
\newblock A survey on motion prediction and risk assessment for intelligent
  vehicles.
\newblock \emph{Robomech Journal}, 1\penalty0 (1):\penalty0 1, 2014.

\bibitem[Naumann et~al.(2019)Naumann, Konigshof, Lauer, and
  Stiller]{naumann2019safe}
Maximilian Naumann, Hendrik Konigshof, Martin Lauer, and Christoph Stiller.
\newblock Safe but not overcautious motion planning under occlusions and
  limited sensor range.
\newblock In \emph{Proc.\ IEEE Intell.\ Veh.\ Symp.}, pages 140--145, 2019.

\bibitem[Orzechowski et~al.(2018)Orzechowski, Meyer, and
  Lauer]{orzechowski2018tackling}
Piotr~F Orzechowski, Annika Meyer, and Martin Lauer.
\newblock Tackling occlusions \& limited sensor range with set-based safety
  verification.
\newblock In \emph{Proc.\ IEEE Intell.\ Trans.\ Syst.\ Conf.}, pages
  1729--1736, 2018.

\bibitem[Papadimitriou and
  Tsitsiklis(1987)]{papadimitriouComplexityMarkovDecision1987}
Christos~H. Papadimitriou and John~N. Tsitsiklis.
\newblock The {{Complexity}} of {{Markov Decision Processes}}.
\newblock \emph{Math.\ of Operations Res.}, 12\penalty0 (3):\penalty0 441--450,
  1987.
\newblock ISSN 0364765X, 15265471.

\bibitem[Press et~al.(1992)Press, Teukolsky, Flannery, and
  Vetterling]{press1992numerical}
William~H Press, Saul~A Teukolsky, Brian~P Flannery, and William~T Vetterling.
\newblock \emph{Numerical Recipes in Fortran 77: The Art of Scientific
  Computing}.
\newblock Cambridge University Press, 1992.

\bibitem[Richter and Wirges(2019)]{Richter2019}
Sven Richter and Sascha Wirges.
\newblock {Fusion of range measurements and semantic estimates in an evidential
  framework}.
\newblock \emph{Technisches Messen}, 86:\penalty0 102--106, 2019.
\newblock \doi{10.1515/teme-2019-0052}.

\bibitem[Shalev-Shwartz et~al.(2017)Shalev-Shwartz, Shammah, and
  Shashua]{shalevshwartz2017formal}
Shai Shalev-Shwartz, Shaked Shammah, and Amnon Shashua.
\newblock On a formal model of safe and scalable self-driving cars, 2017.

\bibitem[Simon(2010)]{simon2010kalman}
Dan Simon.
\newblock Kalman filtering with state constraints: a survey of linear and
  nonlinear algorithms.
\newblock \emph{IET Control Theory \& Appl.}, 4\penalty0 (8):\penalty0
  1303--1318, 2010.

\bibitem[Steyer et~al.(2018)Steyer, Tanzmeister, and Wollherr]{Steyer2018}
Sascha Steyer, Georg Tanzmeister, and Dirk Wollherr.
\newblock {Grid-Based Environment Estimation Using Evidential Mapping and
  Particle Tracking}.
\newblock \emph{IEEE Trans.\ Intell.\ Veh.}, pages 1--1, 2018.
\newblock ISSN 2379-8904.
\newblock \doi{10.1109/TIV.2018.2843130}.

\bibitem[Sun et~al.(2019)Sun, Zhan, Chan, and Tomizuka]{sun2019behavior}
Liting Sun, Wei Zhan, Ching-Yao Chan, and Masayoshi Tomizuka.
\newblock Behavior planning of autonomous cars with social perception.
\newblock In \emph{Proc.\ IEEE Intell.\ Veh.\ Symp.}, pages 207--213, 2019.

\bibitem[Ta{\c s} and Stiller(2018)]{tas2018limited}
{\"O}mer~{\c S}ahin Ta{\c s} and Christoph Stiller.
\newblock {Limited Visibility and Uncertainty Aware Motion Planning for
  Automated Driving}.
\newblock In \emph{Proc.\ IEEE Intell.\ Veh.\ Symp.}, pages 1171--1178, 2018.

\bibitem[Ta{\c s} et~al.(2017)Ta{\c s}, H{\"o}rmann, Schäufele, and
  Kuhnt]{tas2017automated}
{\"O}mer~{\c S}ahin Ta{\c s}, Stefan H{\"o}rmann, Bernd Schäufele, and Florian
  Kuhnt.
\newblock {Automated Vehicle System Architecture with Performance Assessment}.
\newblock In \emph{Proc.\ IEEE Intell.\ Trans.\ Syst.\ Conf.}, pages 1--8,
  2017.

\bibitem[Ta{\c s} et~al.(2018{\natexlab{a}})Ta{\c s}, Hauser, and
  Stiller]{tas2018decision}
{\"O}mer~{\c S}ahin Ta{\c s}, Felix Hauser, and Christoph Stiller.
\newblock {Decision-Time Postponing Motion Planning for Combinatorial Uncertain
  Maneuvering}.
\newblock In \emph{Proc.\ IEEE Intell.\ Trans.\ Syst.\ Conf.}, pages
  2419--2425, 2018{\natexlab{a}}.

\bibitem[Ta{\c s} et~al.(2018{\natexlab{b}})Ta{\c s}, Salscheider, Poggenhans,
  Wirges, Bandera, Zofka, Strauss, Z{\"o}llner, and Stiller]{tas2016making}
{\"O}mer~{\c S}ahin Ta{\c s}, Niels~Ole Salscheider, Fabian Poggenhans, Sascha
  Wirges, Claudio Bandera, Marc~René Zofka, Tobias Strauss, J.~Marius
  Z{\"o}llner, and Christoph Stiller.
\newblock {Making Bertha Cooperate - Team AnnieWAY's Entry to the 2016 Grand
  Cooperative Driving Challenge}.
\newblock \emph{IEEE Trans.\ Intell.\ Transp.\ Syst.}, 19\penalty0
  (4):\penalty0 1262--1276, April 2018{\natexlab{b}}.
\newblock ISSN 1524-9050.
\newblock \doi{10.1109/TITS.2017.2749974}.
\newblock {Date of Publication: 02 October 2017}.

\bibitem[Xu et~al.(2014)Xu, Pan, Wei, and Dolan]{xu2014motion}
Wenda Xu, Jia Pan, Junqing Wei, and John~M Dolan.
\newblock Motion planning under uncertainty for on-road autonomous driving.
\newblock In \emph{Proc.\ IEEE Int.\ Conf.\ Robot.\ and Autom.}, pages
  2507--2512, 2014.

\bibitem[Yin et~al.(2008)Yin, Makris, and Velastin]{yin2008real}
Fei Yin, Dimitrios Makris, and Sergio Velastin.
\newblock Real-time ghost removal for foreground segmentation methods.
\newblock In \emph{IET Electron.\ Lett.}, pages 1351--1353, 2008.
\newblock \doi{10.1049/el:20082118}.

\bibitem[Yu et~al.(2019)Yu, Vasudevan, and Johnson-Roberson]{yu2019occlusion}
Ming-Yuan Yu, Ram Vasudevan, and Matthew Johnson-Roberson.
\newblock Occlusion-aware risk assessment for autonomous driving in urban
  environments.
\newblock \emph{IEEE \ Robot.\ and Autom.\ Lett.}, 4\penalty0 (2):\penalty0
  2235--2241, 2019.

\bibitem[Zhan et~al.(2016)Zhan, Liu, Chan, and Tomizuka]{zhan2016non}
Wei Zhan, Changliu Liu, Ching-Yao Chan, and Masayoshi Tomizuka.
\newblock A non-conservatively defensive strategy for urban autonomous driving.
\newblock In \emph{Proc.\ IEEE Intell.\ Trans.\ Syst.\ Conf.}, pages 459--464,
  2016.

\bibitem[Ziegler et~al.(2014)Ziegler, Bender, Dang, and
  Stiller]{ziegler2014trajectory}
Julius Ziegler, Philipp Bender, Thao Dang, and Christoph Stiller.
\newblock {Trajectory planning for Bertha — a local, continuous method}.
\newblock In \emph{Proc.\ IEEE Intell.\ Veh.\ Symp.}, pages 450--457, 2014.

\end{thebibliography}
\end{document}